\documentclass[conference]{IEEEtran}
\IEEEoverridecommandlockouts
\usepackage{cite}
\usepackage{balance}
\usepackage{changes}
\usepackage{amsmath,amssymb,amsfonts}
\usepackage{algorithmic}
\usepackage{tablefootnote}
\usepackage{graphicx}
\usepackage{multirow}
\usepackage{array}
\usepackage{textcomp}
\usepackage[hidelinks]{hyperref}
\usepackage{url}
\usepackage{changes}
\usepackage{float}
\usepackage[numbers, sort&compress]{natbib}
\usepackage{pifont}
\usepackage[margin=2cm]{geometry} 
\usepackage{caption}

\usepackage{xcolor}
\def\BibTeX{{\rm B\kern-.05em{\sc i\kern-.025em b}\kern-.08em
    T\kern-.1667em\lower.7ex\hbox{E}\kern-.125emX}}

\usepackage{enumitem}
\setlist[itemize]{leftmargin=10pt}
   
\begin{document}

\title{Hybrid ResNet-1D–BiGRU with Multi-Head Attention for Cyberattack Detection in Industrial IoT Environments}

\makeatletter 
\newcommand{\linebreakand}{%
  \end{@IEEEauthorhalign}
  \hfill\mbox{}\par
  \mbox{}\hfill\begin{@IEEEauthorhalign}
}
\makeatother 

\author{\IEEEauthorblockN{Afrah Gueriani}
\IEEEauthorblockA{\textit{LSEA Lab., Faculty of Technology} \\
\textit{University of MEDEA}\\
Medea 26000, Algeria\\
gueriani.afrah@univ-medea.dz }
\and
\IEEEauthorblockN{Hamza Kheddar}
\IEEEauthorblockA{\textit{LSEA Lab., Faculty of Technology} \\
\textit{ University of MEDEA}\\
Medea 26000, Algeria \\
kheddar.hamza@univ-medea.dz}
\and 
\IEEEauthorblockN{Ahmed Cherif Mazari}
\IEEEauthorblockA{\textit{LSEA Lab, Faculty of Science} \\
\textit{ University of MEDEA}\\
Medea 26000, Algeria \\
mazari.ahmedcherif@univ-medea.dz}
}

\makeatletter

\def\ps@headings{%
\def\@oddhead{\parbox[t][\height][t]{\textwidth}{\flushleft

\noindent\makebox[\linewidth]
}
\vspace{0.5cm}
\hfil\hbox{}}%
\def\@oddfoot{\MYfooter}%
\def\@evenfoot{\MYfooter}}
\def\ps@IEEEtitlepagestyle{%

\def\@evenhead{\scriptsize\thepage \hfil \leftmark\mbox{}}%
\def\@oddfoot{979-8-3315-6892-4/25/\$31.00 \textcopyright ©2025   {IEEE} \hfil 
\leftmark\mbox{}}%
\def\@evenfoot{\MYfooter}}
\maketitle

\begin{abstract}
This study introduces a hybrid deep learning model for intrusion detection in Industrial IoT (IIoT) systems, combining ResNet-1D, BiGRU, and Multi-Head Attention (MHA) for effective spatial-temporal feature extraction and attention-based feature weighting. To address class imbalance, SMOTE was applied during training on the Edge-IIoTset dataset. The model achieved 98.71\% accuracy, a loss of 0.0417\%, and low inference latency (0.0001 sec/instance), demonstrating strong real-time capability. To assess generalizability, the model was also tested on the CICIoV2024 dataset, where it reached 99.99\% accuracy and F1-score, with a loss of 0.0028, 0\% FPR, and 0.00014 sec/instance inference time. Across all metrics and datasets, the proposed model outperformed existing methods, confirming its robustness and effectiveness for real-time IoT intrusion detection.

\end{abstract}


\section{Introduction}
\label{sec1}

The rapid advancement and extensive deployment of IoT devices have transformed modern life, enhancing convenience and fostering the development of interconnected systems. However, this technological progress also introduces substantial challenges, particularly the increasing vulnerability of IoT devices to cyber threats \cite{gueriani2024adaptive}. Additionally, the relentless refinement of hacking techniques has given rise to novel and highly sophisticated cyber intrusions, creating complex and unforeseen security challenges \cite{hossain2023ensuring}. The IIoT, which extends IoT to industrial applications, faces even greater risks due to its critical role in sectors like manufacturing, healthcare, and energy. This necessitated the development of advanced and robust security measures, including Network Intrusion Detection Systems (NIDSs) (discussed in \cite{kheddar2024reinforcement}), to identify and alert against attacks and malware at an early stage. Artificial Intelligence (AI), which refers to machines emulating intelligent human behavior \cite{kubat2023fundamentals}, has increasingly emerged as a pivotal component in the domain of network security and intrusion detection \cite{sharma2024explainable}. This study introduces an innovative hybrid framework that integrates ResNet-1D, BiGRU, and Multihead-Attention mechanisms which is discussed in \cite{kheddar2025transformers} to enhance intrusion detection. Additionally, the use of the SMOTE technique addresses data imbalance in EdgeIIoTset dataset, ensuring more accurate classification. 
The proposed ResNet-1D-BiGRU-MHA capture spatial and temporal dependencies in IIoT network traffic. The rationale behind selecting this combination is grounded in their complementary strengths in feature extraction, sequence modeling, and attention-based weighting. ResNet-1D is employed as the initial feature extraction component due to its ability to efficiently capture hierarchical representations in sequential network data. ResNet-1D, in contrast to traditional CNNs, uses residual connections to address the vanishing gradient issue, enabling deeper networks to learn without seeing a drop in performance \cite{zhao2023iot}. Addionaly, BiGRU for temporal dependency modeling is incorporated to model long-term dependencies by reducing computational complexity and allowing the model to learn attack patterns that may depend on both past and future network behavior \cite{drewek2021survey}. However, MHA is integrated to dynamically assign different levels of importance to various features in the time series data. Even while IDS research has advanced significantly, there are still a number of issues that limit the usefulness of current models, especially in IIoT systems. Security teams are overloaded with false alarms due to the high false positive rates (FPR) of many IDS models. Conversely, false negatives pose a critical risk by allowing sophisticated cyberattacks to go undetected. In real-world cybersecurity datasets, attack samples are often much rarer than normal traffic, leading to class imbalance issues. Many existing models fail to effectively learn from minority-class attacks, causing reduced detection accuracy for rare but critical threats like in \cite{zhou2025multi,hernandez2025network}. 

This paper's remaining sections are arranged as follows: The preliminary findings, which include a review of the pertinent background and associated studies, are presented in Section \ref{sec2}. The suggested method and the dataset preprocessing procedures are explained in Section \ref{sec3}. Section \ref{sec4} reports on extensive tests conducted to evaluate the model's performance. A summary of the paper's main conclusions and some avenues for further research are provided in Section \ref{sec5}.

\section{Literature Review}
\label{sec2}

DL approaches have proven highly effective in addressing cybersecurity threats, outperforming traditional centralized machine learning models, particularly in detecting advanced attacks. To further improve performance, researchers are increasingly adopting hybrid and concatenated DL architectures.
Z. Xia et al. \cite{xia2024pso} propose a PSO-GA-optimized ResNet-BiGRU-based intrusion detection method aimed at enhancing network security.
To further improve accuracy, genetic algorithm (GA) and particle swarm optimization (PSO) are employed for hyperparameter tuning. Experiments were demonstrate superior performance on three different dataset namely KDD99, UNSW-NB15, and CIC-IDS 2017 compared to existing methods. 
In \cite{javeed2023explainable}, D. Javeed et al. introduce an explainable and resilient IDS specifically designed for Industry 5.0 environments. 
The proposed model integrates BiLSTM and BiGRU architectures to enhance detection accuracy. Using the CICDDoS2019 dataset, the system effectively detects and eliminates cyber-threats in interconnected industrial systems. 
In \cite{xiang2024resnest}, Y. Xiang et al. propose a hybrid DL model combining ResNet and biGRU for effective intrusion detection in IoT environments. 
Numerous experiments show that the model performs better than current techniques, achieving better detection rates and robustness on  three benchmark IoT datasets namely: NBaIoT, PreIoT, and UNSW-NB15. In \cite{gueriani2025cyber}, the authors investigated the feasibility of developing a DNN-GRU model enhanced with Multi-Head Attention. The experimental results demonstrated the superiority of the proposed model, achieving accuracy rates of 98.22\% and 99.78\% on medical and industrial datasets, respectively. In their subsequent works \cite{gueriani2025explainable,gueriani2025robust}, the same authors employed SHAP with multiple GRU-BiLSTM-MHA-based IDS combinations to analyze and rank feature importance, with the aim of reducing computational cost. Evaluated on the same datasets before and after the application of balancing techniques, the proposed architecture achieved high accuracy across domains, demonstrating strong generalization capability and adaptability under both settings. A Vision Transformer-BiLSTM architecture was also investigated in \cite{gueriani2025se} for the development of an advanced IDS method. The experimental results demonstrated that the proposed ViT-BiLSTM model outperformed many existing approaches across multiple evaluation metrics.

\section{Proposed Methodology} 
\label{sec3}
This section presents a background of attention mechanism, then discuss the suggested model, a ResNet-1D-BiGRU-MHA hybrid, will take up this section. This architecture is created to recognize and classify benign and dangerous traffic in a dataset of novel environments.

\noindent \textbf{\textit{- Attention Mechanism:}} The self-attention mechanism is a powerful and highly efficient approach commonly used in modern deep learning architectures, especially in models that process sequential data such as text, audio, or time series. Three vectors; the query, key, and value vectors; represent each component of the input sequence in this technique. Throughout the training process, these vectors are discovered and improved. The core idea is to determine the degree of attention an element (represented by the query) should allocate to another element (represented by the key) by computing a compatibility score, which guides the weighted aggregation of the corresponding value vectors. In order to ensure that the scores total up to one, this score is usually calculated by taking the dot product of the query and key vectors, then adding a scaling factor and performing a softmax operation \cite{vaswani2017attention}. 

See Equation 1 in in \cite{qathrady2024sacnn} for a thorough description of the attention mechanism:

\begin{equation}
    \small
    \mathrm{Attention (Q, K, V)}= softmax\left(\frac{Q \times K^T}{\sqrt{d_{q}}}\right) \times V
    \label{eq11}
\end{equation}

Where:

\begin{itemize}
\item \textit{Q (Query)} represents the transformed input used by the model to compute attention scores. As indicated by Equation \ref{eq2}, these vectors are calculated by multiplying the weight matrix \( W_Q \) by the input \( X \)  and correspond to various segments of the input sequence:

    \begin{equation}
    \small
        \mathrm{Q}= X \times W_{Q}
        \label{eq2}
    \end{equation}

\item \textit{K (Key)} refers to the set of vectors against which the query is compared to determine relevance. These vectors correspond to various input sequence segments and are computed by multiplying the input \( X \) with the weight matrix \( W_K \), according to \ref{eq3}:
    \begin{equation}
    \small
        \mathrm{K}= X \times W_{K}
        \label{eq3}
    \end{equation}

\item \textit{V (Values)} are the values associated with each key. These vectors hold the actual information the model will utilize after computing the attention scores, as given by the following equation:
    \begin{equation}
    \small
        \mathrm{V}= X \times W_{V}
        \label{eq4}
    \end{equation}

\item \textit{d\_{q}} is the dimensionality of the query vector Q.
\end{itemize}

\subsection{Data Preparation}
Effective data preparation enhances the reliability and accuracy of the proposed models. 
The following points detail the key steps for data transformation:

\noindent \textbf{\textit{- Data preprocessing:}} The initial step for both datasets involved transforming the feature sets; comprising 60 features and 15 distinct attack types in the Edge-IIoT dataset, and 9 features with 6 attack types in the second dataset; into a two-dimensional format to align with the input requirements of the proposed model.

\noindent \textbf{\textit{- Addressing class imbalance with SMOTE:}} a popular oversampling technique that creates synthetic instances for the minority class using interpolation from the available samples, was utilized in the study to correct class imbalance \cite{sayegh2024enhanced}. 

\noindent \textbf{\textit{- Data splitting:}} Following preprocessing, a total of 60 features from the Edge-IIoT dataset and 9 features from the CICIoV2024 dataset were retained for model training. Each dataset was subsequently partitioned into training and testing subsets using an 80/20 split, respectively.
During the training phase, the Adam optimization technique was used to facilitate effective convergence and raise the model's overall efficacy.

\noindent \textbf{\textit{- Data Numerization:}} To facilitate the processing of non-numeric categorical data, a numerization approach was employed. Specifically, categorical features were converted into numerical representations utilizing the LabelEncoder in conjunction with the \textbf{fit\_transform} method.

\subsection{Model architecture}
The proposed model combines ResNet-based feature extraction, BiGRU for sequence modeling, and MHA (Figure \ref{fig40}) to enhance intrusion detection in IIoT environments. This hybrid approach captures spatial and temporal dependencies while improving classification accuracy. The proposed architecture begins with the \textit{\textbf{Input Layer:}} The model accepts inputs of shape (60, 1), where 1 denotes a single channel per feature and 60 denotes the number of features. \textbf{\textit{ResNet Block:}} At the beginning of the model, a residual connection with two convolutional layers (Conv1D) is employed for feature extraction. The first Conv1D layer uses 64 filters with a kernel size of 3, followed by BatchNormalization for stabilization. 
Another Conv1D layer with the same configuration is applied, and a residual connection adds a \textbf{Conv1D(1x1)} layer to match dimensions. 
 \textit{\textbf{BiGRU}} layer with 64 units is applied to capture sequential patterns in the data. The BiGRU processes the output from the convolutional layers, leveraging both forward and backward information from the sequence. \textit{\textbf{LayerNormalization}} is included to stabilize the output from the GRU layer. A \textit{\textbf{MHA}} mechanism with \texttt{num\_heads=4} and \texttt{key\_dim=64} is subsequently applied to the output of the BiGRU layer.  After that, a \textit{\textbf{Dropout}} layer is used at a rate of 0.5 to randomly deactivate neurons during training to reduce overfitting. The attention mechanism's output is compressed into a one-dimensional vector before being sent to the fully connected layers for categorization. For additional processing and abstraction, two \textbf{\textit{Dense}} layers are employed, each having "ReLU" activation with "64 units" and "32 units", respectively. The six different classes in the multiclass classification task are represented by the "6 units" with a "softmax activation function" in the final output layer.

\begin{figure}[h!]
\centering
\includegraphics[scale=0.35]{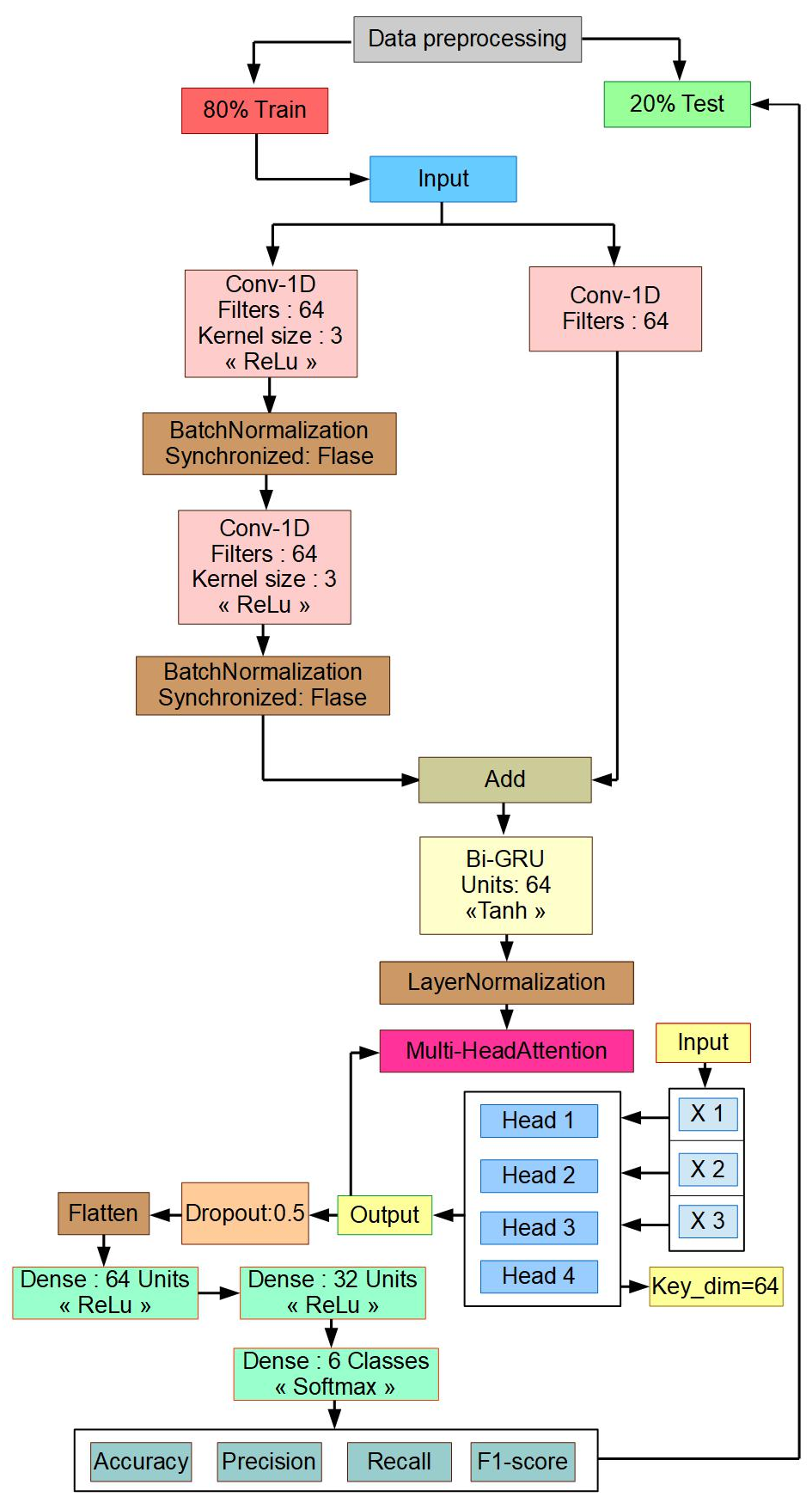}
\caption{The proposed framework architecture.}
\label{fig40}
\end{figure}

\section{Experimentation, Outcomes and discussion}
\label{sec4}

\subsection{Dataset description}
\noindent \textbf{- Edge-IIoTset dataset}\footnote{\url{ https://www.kaggle.com/datasets/cnrieiit/mqttset}}: Emerged as a widely recognized benchmark within the research community for evaluating AI-based IDSs, particularly in real-time applications. This dataset encompasses IoT and IIoT traffic data gathered from a real-world testbed featuring seven interconnected layers and ten smart devices and sensors \cite{ferrag2022edge}. It includes 15 traffic types,  
grouped into six distinct categories. Initially comprising 1,176 features, the dataset was refined to 61 relevant features focusing on IoT devices. \\ 
\noindent {\textbf{- CICIoV2024}}\footnote{\url{ https://www.unb.ca/cic/datasets/iov-dataset-2024.html}}: Introduced for intrusion detection in the context of the Internet of Vehicles (IoV). It contains five range of modern attack types in addition to normal traffic. The dataset was captured in a realistic testbed environment simulating in-vehicle and vehicular communication scenarios. The dataset comprises three distinct representations; binary, decimal, and hexadecimal; each characterized by a different set of features \cite{carlos2024ciciov2024}.

\subsection{Performance metrics}
To evaluate how well the suggested model detects different kinds of attacks, standard assessment metrics.  
The four basic classification outcomes; true positive (TP), true negative (TN), false positive (FP), and false negative (FN); are the source of these measures, as described in \cite{gueriani2023deep, kheddar2024deep, kheddarASR2023, gueriani2025adaptive}. In this case, FP and FN stand for the misclassified legitimate and attack occurrences, respectively, whereas TP and TN indicate the number of correctly classified attack and valid cases, respectively \cite{dunn2020robustness}. The corresponding equations for each metric are presented below:

    \begin{equation*}
    \small
        \mathrm{Acc}=\mathrm{\frac{TP_{A}+TN_{N}}{TP_{A}+FP_{N}+TN_{N}+FN_{A}}},\mathrm{Rc}=\mathrm{\frac{TP_{A}}{TP_{A}+FN_{A}}} 
        \label{eq5}
    \end{equation*}

    \begin{equation*}
    \small
         \mathrm{Pr}=\mathrm{\frac{TP_{A}}{TP_{A}+FP_{N}}}, \hspace{0.2cm} \mathrm{F1-Score}= \mathrm{2 \times \frac{Pr \times Rc}{Pr + Rc}}
        \label{eq7}
    \end{equation*}

        \begin{equation*}
    \small
    \mathrm{FPR}= \mathrm{\frac{FP_{N}}{FP_{N}+TN_{N}}}
        \label{eq9}
    \end{equation*}

Where the indices \( \mathrm{A} \) and \( \mathrm{N} \) refer to abnormal and normal samples, respectively.

\subsection{Experiments and Results}
The experimental findings of assessing the suggested model using a variety of performance measures on the Edge-IIoTset dataset are shown in this section. The training was conducted in the Kaggle environment utilizing a dual GPU T4 setup with 15 GiB of memory, using the Adam optimizer for fifteen epochs.

\begin{itemize}

\item \textbf{\textit{{Accuracy and loss graph:}}} Figure \ref{fig4} Figure 2 (a) and (b) display the training/validation accuracy and loss of the proposed model. The validation accuracy reaches 98.71\% and closely follows the training accuracy, suggesting good generalization without overfitting. Both losses stabilize early and remain low, with the validation loss settling at 0.0417, indicating effective model optimization.

\begin{figure}[h!]
\centering
\includegraphics[scale=0.4]{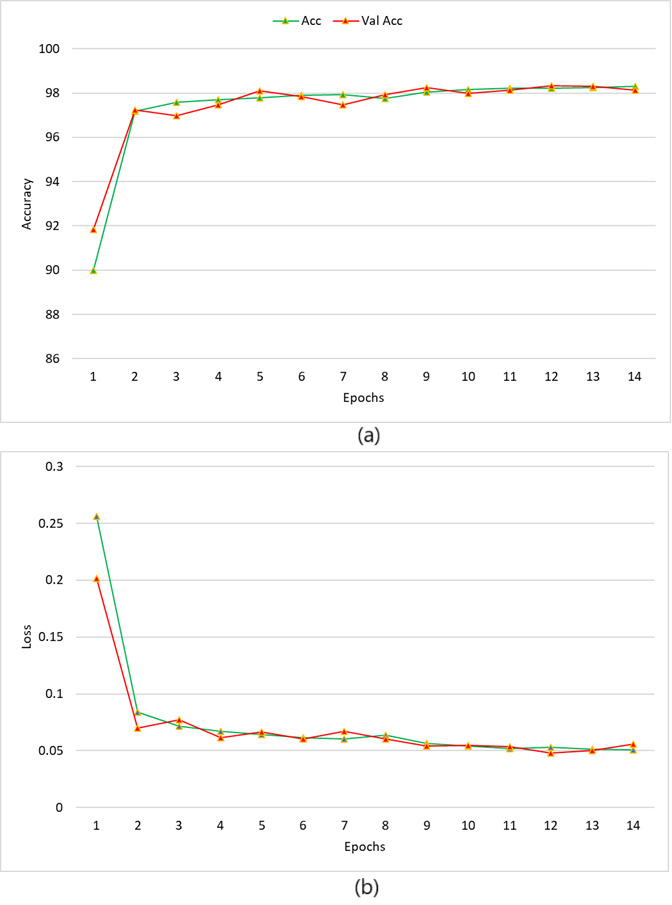}
\caption{Accuracy and loss of the proposed ResNet-1D-BiGRU-MHA model. (a): Accuracy graph, (b): Loss graph}
\label{fig4}
\end{figure} 

\end{itemize}

\begin{itemize}
\item \textbf{\textit{{Classification Report:}}}
Table \ref{Mc} shows the proposed model's strong performance in multiclass attack detection. It achieves perfect accuracy for normal traffic, information gathering, MITM, and malware attacks, and performs well on DDoS and injection attacks with minimal misclassifications. Overall, the ResNet-1D-BiGRU-MHA architecture demonstrates effective classification capability and is well- for real-time cybersecurity threat detection.

\end{itemize}

\begin{center}
\begin{table}
\centering
\caption{Multiclass classification report of the proposed ResNet-1D-BiGRU-MHA approach.}
\begin{tabular}{llllllll}
\hline
& Precision & Recall & F1 & Support \\ [0.6mm]
\hline
Benign traffic (0) & 100\% & 100\% & 100\% & 9860 \\[0.6mm]

DDoS (1) & 99\% & 97\% & 98\% & 10016\\ [0.6mm]

Info. gathering (2) & 100\% & 100\% & 100\% & 9878 \\[0.6mm]

MITM (3) & 100\% & 100\% & 100\% & 9815\\ [0.6mm]

Injection (4) & 97\% & 99\% & 98\% & 9779 \\[0.6mm]

Malware (5) & 100\% & 100\% & 100\% & 9928\\ [0.6mm]
\hline

\hline
\end{tabular}
\label{Mc}
\end{table}
\end{center}

\begin{itemize}
\item \textbf{\textit{{Confusion matrix:}}}
Figure \ref{fig7} shows the confusion matrix for the ResNet-1D-BiGRU-MHA model, demonstrating strong classification performance. Most attack types are correctly identified, with high accuracy along the diagonal and minimal misclassifications. Notably, only 4\% of "malware" attacks are misclassified as "DDoS", and 2\% of "DDoS" attacks as "malware", indicating effective differentiation between attack types.

\begin{figure}[h!]
\centering
\includegraphics[scale=0.2]{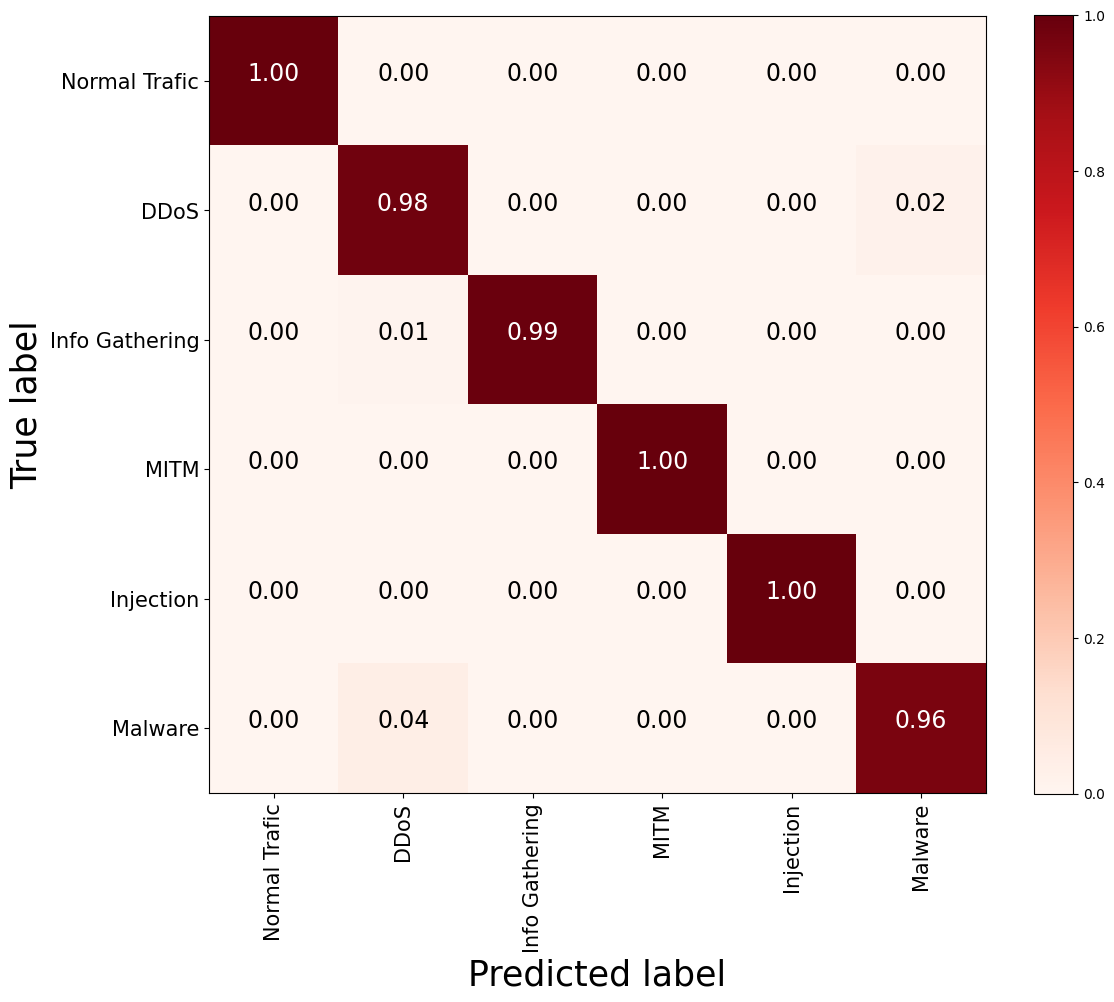}
\caption{Confusion matrix of the proposed ResNet-1D-BiGRU-MHA model}
\label{fig7}
\end{figure} 

\end{itemize}

\begin{itemize}
    \item \textbf{\textit{Inference and training time:}} In real-time IDS, the primary emphasis is placed on inference time rather than training duration. In this study, the ResNet-1D-BiGRU-MHA model achieves an inference time of 0.0001 seconds per instance, while the ResNet-1D-BiGRU-MHA model requires 58.07 seconds per epoch for training. These results highlight the exceptionally fast performance of the proposed models, particularly during inference.

\end{itemize}

\begin{itemize}

\item \textbf{\textit{FPR:}} A value of 0.002\% in multiclass attack detection indicates an exceptionally low rate of false alarms, proving the accuracy of the model in differentiating between harmful and benign traffic.

\end{itemize}

\begin{itemize}

\item \textbf{\textit{{Receiver operating characteristics (ROC):}}} Figure \ref{fig9} presents the ROC curves for the proposed multiclass attack detection model. The ROC curves for all classes (0 to 5) are positioned at the top-left corner of the plot, indicating ideal classification performance. An AUC value of 1.00 across all classes further confirms that the model achieves perfect accuracy in distinguishing true positives and true negatives for each attack category.

\begin{figure}[h!]
\centering
\includegraphics[scale=0.4]{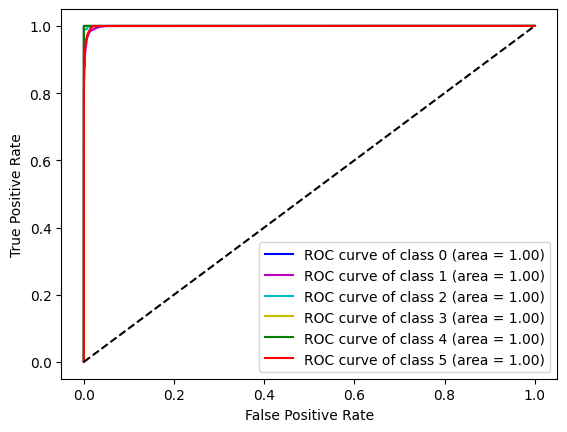}
\caption{ROC curves for multiclass classification of the proposed ResNet-1D-
BiGRU-MHA model.}
\label{fig9}
\end{figure}

\end{itemize}

\begin{itemize}

\item \textbf{\textit{{Performance Evaluation Against Existing Methods:}}}
Table \ref{tab4} presents a comparative analysis of the proposed ResNet-1D-BiGRU-MHA model against recent state-of-the-art approaches on the Edge-IIoTSet and CICIoV2024 datasets. The proposed model consistently outperforms existing methods across all major evaluation metrics.

On Edge-IIoTSet, it achieves an accuracy of 98.71\%, surpassing models like BiGRU-LSTM \cite{javeed2023intrusion}. Although LSTM-CNN-Att \cite{gueriani2025adaptive} slightly exceeds this with 99.04\% accuracy, it was tested on a single dataset and lacks inference time reporting, limiting its practical evaluation. In contrast, the proposed model demonstrates high performance and low inference latency (0.0001 sec/instance) across both datasets, confirming its generalizability and real-time suitability.

On CICIoV2024, the model achieves 99.99\% accuracy, with near-perfect precision, recall, F1-score, and 0.0000\% FPR, significantly outperforming models such as CNN-LSTM-ViT \cite{jailani2025hybrid} and XGB \cite{ccolhak2025accelerating}. Its inference time on this dataset is also the lowest (0.00014 sec/instance) compared to 0.0213 sec for CNN-LSTM-ViT, highlighting its computational efficiency.

\begin{table*}
\centering
\caption{Comparing the Best Practices for Multiclass Classification on the Edge-IIoTset Dataset with Performance Metrics for the Suggested ResNet-1D-BiGRU-MHA Model.}
\label{tab4}
\begin{tabular}{|c|c|c|c|c|c|c|c|c|c|}
\hline
Work  & Model  & Dataset & Acc (\%) & Loss & Pr (\%) & Rc (\%) & F1 (\%) & FPR (\%)& Inf time (Sec/Inst)\\
\hline

\cite{gueriani2025adaptive} & LSTM-CNN-Att & EdgeIIoT & 99.04 & 0.0220  & 99.05 & 99.04 & 99.04 & 0.002 & \ding{55}\\[0.6mm]

\cite{carlos2024ciciov2024} & DNN & CICIoV2024 & 96 & \ding{55}  & 83 & 76 & 78 & \ding{55}& \ding{55}\\[0.6mm]

\cite{javeed2023intrusion} & BiGRU-LSTM & EdgeIIoT & 98.32 & \ding{55}  & 98.78 & 97.22 & \ding{55} & \ding{55} & \ding{55}\\[0.6mm]

\cite{jailani2025hybrid} & CNN-LSTM-ViT & CICIoV2024 & 99.78 & \ding{55}  & \ding{55} & \ding{55} & 99.65 & 1.2 & 0.0213 \\[0.6mm]

\hline
\hline

\textbf{Presented} & \textbf{ResNet-1D-BiGRU-MHA} & \textbf{EdgeIIoT} & \textbf{98.71} & \textbf{0.0417} & \textbf{98.71} & \textbf{98.70} & \textbf{98.71} & \textbf{0.002}& \textbf{0.0001}\\

&  & \textbf{CICIoV2024} & \textbf{99.99} & \textbf{0.0028} & \textbf{99.99} & \textbf{99.99} & \textbf{99.99} & \textbf{0.0000}& \textbf{0.00014}\\

\hline
\end{tabular}
\end{table*}

\end{itemize}

\begin{itemize}
    \item \textbf{\textit{{Ablation study:}}}
To evaluate the contribution of individual architectural components to the overall performance of the proposed model, an ablation study was conducted. Various model configurations were analyzed, with the results summarized in Table III \ref{tab5}. The analyzed configurations include: ResNet-1D, BiGRU-MHA, ResNet-1D-BiGRU, and ResNet-1D-BiGRU-MHA, with varying numbers of attention heads, dropout rates, and dense layers.

\begin{itemize}
\item The baseline model (ResNet-1D, \textbf{\#1}) showed limited performance (47.95\% accuracy), highlighting the absence of temporal feature extraction. 

\item Adding BiGRU with MHA (\textbf{\#2}) improved accuracy, though the loss remained high, indicating further enhancement was needed. 

\item Model \#3 combined ResNet-1D and BiGRU, improving temporal learning, but with loss values similar to \#2, suggesting attention mechanisms could enhance performance.

\item The proposed full model (\textbf{\#5}), integrating ResNet-1D, BiGRU, and MHA (with four attention heads and 0.5 dropout), achieved the best results; 98.71\% accuracy, low FPR, and minimal inference time.

\item Varying attention heads showed that reducing heads to two (\textbf{\#4}) maintained high performance, while increasing to eight (\textbf{\#6}) led to decreased accuracy, suggesting an optimal head count.

\item Dropout variations in models \textbf{\#7} (0.3) and \textbf{\#8 }(0.7) had minimal effect, indicating robustness to dropout rate changes in this range.

\item Model \#10, identical to \#5 but without SMOTE, showed degraded performance, underlining the importance of class imbalance handling.

\end{itemize}

This analysis confirms the synergistic effect of combining spatial (ResNet-1D), temporal (BiGRU), and attention mechanisms (MHA), along with balanced data processing, in achieving optimal model performance.

\begin{table*}[ht!]
\centering
\caption{Performance of different variants of the proposed models in multiclass classification.}
\scriptsize
\label{tab5}
\begin{tabular}{|c |c| c| c|c| c| c| c|}

\hline

 Case number & Model& N. of Att heads & Dropout (\%) & Accuracy (\%)& Loss(\%) & FPR(\%)& Inf. time\\
\hline

\#1 & ResNet-1D &\ding{55} & 0.5&47.95 &2.7668 & 0.0130&0.00008\\

\#2 & BiGRU-MHA &4 & 0.5&98.28&0.0560 & 0.0034&0.0002\\

\#3 & ResNet-1D-BiGRU &\ding{55} & 0.5&98.07&0.0557 & 0.0039&0.0003\\

\#4  & ResNet-1D-BiGRU-MHA & 2& 0.5 & 98.27 & 0.0504&0.0033 &0.00015\\

\textbf{\#5 }& \textbf{ResNet-1D-BiGRU-MHA}& \textbf{4}& 0.5& \textbf{98.71}& \textbf{0.0294}&\textbf{0.0020}&\textbf{0.0001}\\

\#6 & ResNet-1D-BiGRU-MHA &8 & 0.5 & 97.97 & 0.0613& 0.0038&0.00016\\

\#7 & ResNet-1D-BiGRU-MHA &4 & 0.3 & 97.95 & 0.0650& 0.0025&0.00017\\

\#8 & ResNet-1D-BiGRU-MHA &4 & 0.7 & 98.38 & 0.0524& 0.0033&0.00017\\

\#9 & ResNet-1D-BiGRU-MHA 
 &4 & 0.5 & 98.30 & 0.0500& 0.0033&0.00017\\
& with only 2 dense layers & &  &  & & &\\

\#10 & ResNet-1D-BiGRU-MHA 
 &4 & 0.5 & 96.97 & 0.0873& 0.0053&0.00015\\
& without SMOTE technique & &  &  & & &\\

\hline
\end{tabular}
\end{table*}

\end{itemize}

\section{Conclusion}
\label{sec5}

The proposed model combines SMOTE, MHA, and a hybrid ResNet-1D-BiGRU architecture to effectively detect cyberattacks in IIoT environments. Validated on the Edge-IIoTset dataset, the model achieves over 98\% accuracy, high F1-scores, low FPR, and minimal inference times; making it highly appropriate for real-time intrusion detection.

A detailed ablation study confirms the contribution of each architectural component to overall performance. Further evaluation on the CICIoV2024 dataset shows the model maintains high accuracy (over 99\%), outperforming previous approaches and demonstrating strong generalization.

While the results are promising, future work should explore broader dataset evaluations and address evolving threats. Potential directions include incorporating transfer learning for adaptability, reinforcement learning for hyperparameter tuning, and Explainable AI (XAI) to enhance interpretability and reliability of the system.

\section*{Acknowledgment}

The authors acknowledge that the study was partially funded by the PRFU-A25N01UN260120230001 grant from the Algerian Ministry of Higher Education and Scientific Research.

\balance
\bibliographystyle{IEEEtran}
\begin{footnotesize}
\bibliography{references.bib}
\end{footnotesize}
\end{document}